\documentclass[runningheads]{llncs}
\usepackage{graphicx}

\usepackage{tikz}
\usepackage{comment}
\usepackage{amsmath,amssymb} 
\usepackage{color}

\DeclareMathOperator*{\argmin}{\arg\!\min}

\usepackage[accsupp]{axessibility}  

\usepackage{caption}
\usepackage{subcaption}

\begin{document}
\pagestyle{headings}
\mainmatter
\def\ECCVSubNumber{100}  

\title{qDWI-Morph: Motion-compensated quantitative Diffusion-Weighted MRI analysis for fetal lung maturity assessment\thanks{This research was supported in part by a grant from the United States-Israel Binational Science Foundation (BSF), Jerusalem, Israel.}
} 


\titlerunning{qDWI-Morph: Motion-compensated quantitative DWI analysis}
%
\author{Yael Zaffrani-Reznikov\inst{1}\orcidID{0000-0003-3817-2433} \and
Onur Afacan\inst{2}\orcidID{0000-0003-2112-3205} \and
Sila Kurugol\inst{2}\orcidID{0000-0002-5081-4569} \and
Simon Warfield\inst{2}\orcidID{0000-0002-7659-3880} \and 
Moti Freiman\inst{1}\orcidID{0000-0003-1083-1548}}
\authorrunning{Y. Zaffrani-Reznikov et al.}
%
\institute{Faculty of Biomedical Engineering, Technion, Haifa, Israel \and
Boston Children's Hospital, Boston, MA, USA 
\\
\email{yael.rez@campus.technion.ac.il}
}

\maketitle

\begin{abstract}
Quantitative analysis of fetal lung Diffusion-Weighted MRI (DWI) data shows potential in providing quantitative imaging biomarkers that indirectly reflect fetal lung maturation. However, fetal motion during the acquisition hampered quantitative analysis of the acquired DWI data and, consequently, reliable clinical utilization. We introduce qDWI-morph, an unsupervised deep-neural-network architecture for motion compensated quantitative DWI (qDWI) analysis. Our approach couples a registration sub-network with a quantitative DWI model fitting sub-network. We simultaneously estimate the qDWI parameters and the motion model by minimizing a bio-physically-informed loss function integrating a registration loss and a model fitting quality loss. 
We demonstrated the added-value of qDWI-morph over: 1) a baseline qDWI analysis without motion compensation and 2) a baseline deep-learning model incorporating registration loss solely. The qDWI-morph achieved a substantially improved correlation with the gestational age through in-vivo qDWI analysis of fetal lung DWI data ($R^2=0.32$ vs. $0.13$, $0.28$).
Our qDWI-morph has the potential to enable motion-compensated quantitative analysis of DWI data and to provide clinically feasible bio-markers for non-invasive fetal lung maturity assessment. Our code is available at: https://github.com/TechnionComputationalMRILab/qDWI-Morph. 

\end{abstract}

\section{Introduction}
Fetal lung parenchyma maldevelopment may lead to life-threatening physiologic dysfunction due to pulmonary hypoplasia and pulmonary hypertension \cite{lakshminrusimha2015persistent}. Accurate assessment of lung maturation before delivery is critical in determining pre-natal and post-natal care and potential interventions  
 \cite{ahlfeld2014assessment}.
In current clinical practice, the non-invasive assessment of fetal lung parenchyma development relies upon fetal lung volume estimation from either ultrasonography \cite{Moeglin2005FetalUltrasound}, or anatomical magnetic resonance imaging (MRI) \cite{deshmukh2010mr} data. However, these modalities are limited in providing an insight into lung function and are therefore suboptimal in assessing fetal lung maturity and in providing relevant biomarkers for fetal lung parenchyma maldevelopment.

Diffusion-weighted MRI (DWI) is a non-invasive imaging technique sensitive to the random movement of individual water molecules. The displacement of individual water molecules results in signal attenuation in the presence of a magnetic field encoding gradient pulses. This attenuation increases with the degree of sensitization-to-diffusion of the MRI pulse sequence (the “b-value”) \cite{Afacan2016FetalAge}. Quantitative analysis of the DWI data (qDWI) has been suggested previously for functional assessment of fetal lung parenchyma development \cite{moore2001vivo,Afacan2016FetalAge}. 
The qDWI analysis is performed by a least-squares fitting of a signal decay model describing the DWI signal attenuation as a function of the b-value to the acquired DWI data. Commonly, a mono-exponential signal decay model of the form of:  
\begin{align}
  S_{i} = {S_{0}}\cdot e^{-b_i\cdot ADC} \;
  \label{eq:MonoModel}
\end{align}
Where $S_i$ is the signal at b-value $b_i$, $S_0$ is the signal without sensitizing the diffusion gradients, and $ADC$ is the apparent diffusion coefficient, which represents the overall diffusivity, is used in clinical practice \cite{koh2007diffusion}. 

However, qDWI analysis is intrinsically highly susceptible to gross motion between the acquisition of the different b-value images \cite{Afacan2016FetalAge}. Hence, the irregular and unpredictable motion of the fetus, in addition to the maternal respiratory and abdominal motion, causes misalignment between image volumes, acquired at multiple b-values, and impairs the accuracy and robustness of the signal decay parameter estimation \cite{Kurugol2017Motion-robustEstimation}.

Specifically, Afacan et al. \cite{Afacan2016FetalAge} examined the influence of the presence of motion during the DWI data acquisition on the correlation between the overall diffusivity characterized by the ADC parameter and the gestational age (GA). They found that the dependence of the ADC parameter on the GA is of an exponential saturation type, with an R-squared $(R^2)$ value of 0.71 for fetal DWI data that was not affected by fetus motion. However, when fetal DWI data with gentle motion were included, the $R^2$ dropped to 0.232, and when fetal DWI data with severe motion were included, the $R^2$ dropped to 0.102.

We introduce qDWI-Morph, an unsupervised deep-neural-network model for motion-compensated qDWI analysis. Our specific contributions are: 1) unsupervised deep-neural-network (DNN) model for simultaneous motion compensation and qDWI analysis, 2) Introduction of a bio-physically-informed loss function incorporating registration loss with qDWI model fitting loss, and 3) improved correlation between qDWI analysis by means of the ADC parameter and GA demonstrated using in-vivo DWI data of fetal lungs.

\subsection{Background}
\subsubsection{Fetal lung development}

The normal fetal lung parenchyma development starts in the second trimester and progresses through multiple phases before becoming fully functional at full term. The first phase of development, the embryonic and pseudo glandular stage, is followed by the canalicular phase, which starts at 16 weeks, then the saccular stage, which starts at 24 weeks, and ends with the alveolar stage, which starts at 36 weeks gestation. \cite{schittny2017development}. The formation of a dense capillary network and a progressive increase in pulmonary blood flow, leading to reduced extra-cellular space on the one hand and increased perfusion on the other hand, characterizes the progression through these phases \cite{Ercolani2021IntraVoxelMaturation}. Hence, changes in the overall tissue diffusivity are thought to serve as a functional indicator of lung development.  

\subsubsection{Diffusion-weighted MRI}

DWI is a non-invasive imaging technique sensitive to the random movement of individual water molecules. The movement of water molecules depends on tissue micro-environments and results from two phenomena: diffusion and perfusion \cite{Ercolani2021IntraVoxelMaturation}. The water molecule mobility attenuates the DWI signal according to the degree of sensitivity-to-diffusion, the "b-value", used in the acquisition. Typically, DWI images are acquired at multiple b-values, and the signal decay rate parameters are quantified per voxel by least-squares model fitting  \cite{Kurugol2017Motion-robustEstimation}. DWI  can potentially be a functional biomarker for lung maturity since it results from diffusion and perfusion.
According to the mono-exponential model, the signal of a particular voxel decreases exponentially with an increasing b-value (Eq.~\ref{eq:MonoModel}), with a decay rate that depends on the overall diffusivity of the voxel, the ADC, encapsulating both pure diffusion and perfusion influence on the DWI signal. Quantifying the ADC out of the measurements is done by applying $log(\cdot)$ on both sides of Eq.~\ref{eq:MonoModel}:
\begin{align}
  log(S_{i}) = log(S_{0})-b_i\cdot ADC \;
  \label{eq:MonoModelLog}
\end{align}

A set of B equations is obtained by scanning with B different b-values. This linear set of equations can be solved by linear least-squares regression (LLS) \cite{LinearLeastSquares}. The solution is given by:

\begin{align}
\hat{x}_{LS}  & = \argmin_{S_0,ADC}\sum_{i\in B}|\log{(S_i)}-\log{(S_0)}+b_{i}ADC|^2 \; \\
& = (A^TA)^{-1}A^T\beta
\label{eq:LLS}
\end{align}

where: 
\begin{align}
A = \begin{pmatrix}
1 & -b_0\\ 
\vdots  & \vdots\\ 
1 & -b_B
\end{pmatrix},
&& \beta = \begin{bmatrix}
log(S_0), & \cdots,  & log(S_B) 
\end{bmatrix}^T,
&& x = \begin{bmatrix}
log(S_0), & ADC 
\end{bmatrix}^T
\end{align}

In order to be stable against outliers in the acquired DWI signals, a more robust solution can be obtained by using an iterative-least squares algorithm \cite{Burrus2018IterativeRL}. This is an iterative method in which each step involves solving a weighted least squares problem:
\begin{align}
\hat{x}_{IRLS}^{(t)} & = \argmin_{S_0,ADC}\sum_{i\in B}w_i^{(t)}|\log{(S_i)}-\log{(S_0)}+b_{i}ADC|\\
& = (A^TW^{(t)}A)^{-1}A^TW^{(t)}b
\label{eq:IRLS}
\end{align}

where $W^{(t)}$ is a diagonal matrix of weights, with all elements set initially to:
\begin{align}
\left \{w_i^{(0)}\right \}_{i\in B} =  1
\end{align}

and updated after each iteration to:
\begin{align}
\left \{w_i^{(t)}\right \}_{i \in B} =  \frac{1}{\max \left \{|Ax^{(t-1)}-b|_i, 0.0001 \right \}} 
\end{align}

\begin{figure}[t]
	\centering
    \includegraphics[height=6cm]{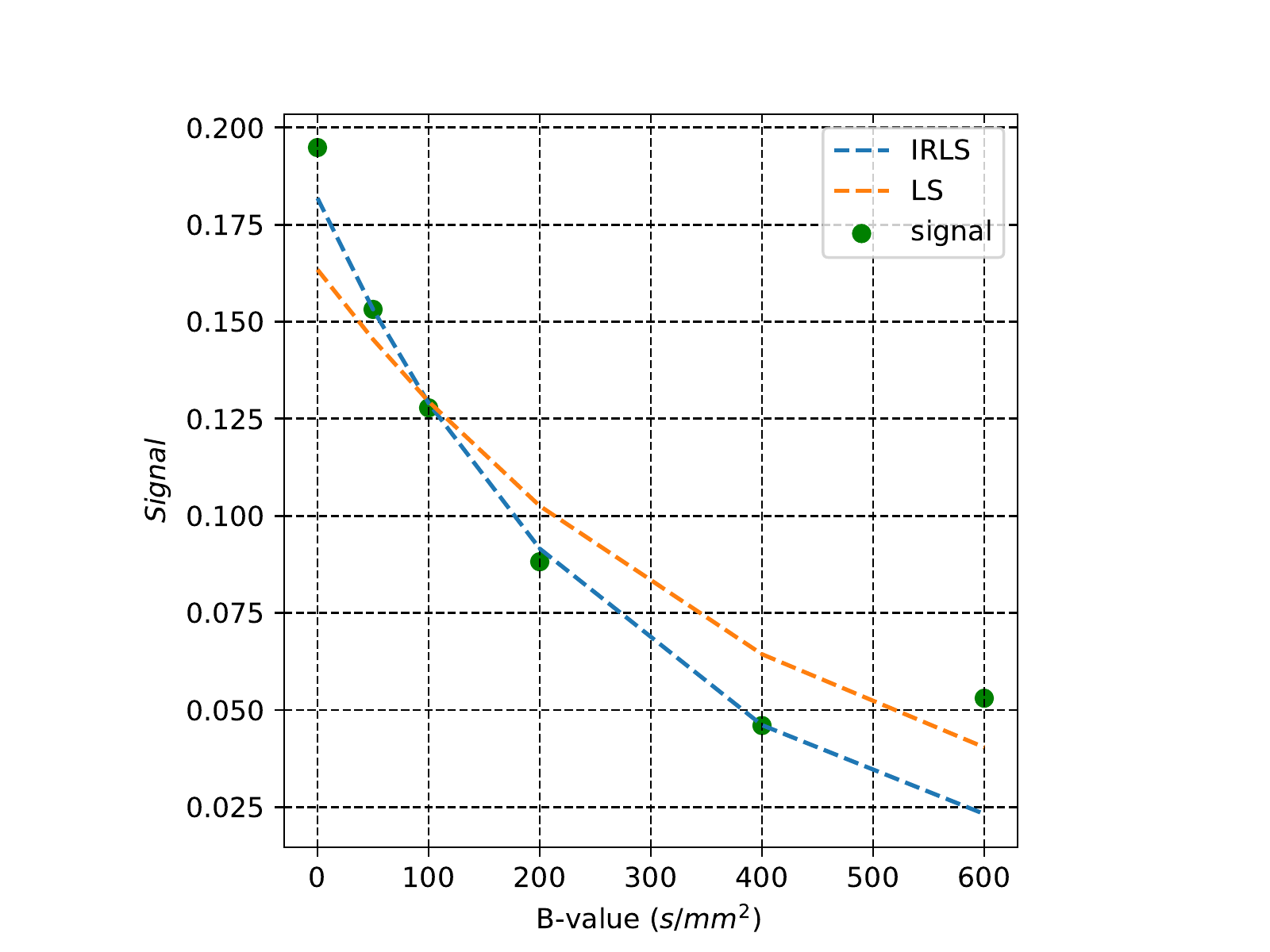}
    \caption{LLS Vs. IRLS fitting of DWI Signal.}
	\label{fig:LS_IRLS}
\end{figure}

As shown In Fig.~\ref{fig:LS_IRLS}, the IRLS is more accurate and ignores outliers by setting them a low weight. On the other hand, the IRLS algorithm is computationally expensive and time-consuming.

\subsection{Fetal Lung Maturity Assessment with qDWI}
Assessing fetal lung maturation using qDWI was first suggested by Moore et al. \cite{moore2001vivo} in 2001. More recently, Afacan et al. \cite{Afacan2016FetalAge} demonstrated a strong association between ADC and GA in fetuses with normal lung development. The ADC is quantified per voxel by LLS regression (Eq.~\ref{eq:LLS}). Then, the averaged ADC in the lung can indicate the GA of the fetus according to the saturation-exponential model suggested by \cite{Afacan2016FetalAge}:
\begin{align}
  ADC = ADC_{sat}(1-e^{-\alpha \cdot GA}) \;
  \label{eq:GA_ADC}
\end{align}

Since the ADC quantification is done voxel-wise, it is sensitive to potential misalignment between image volumes acquired at different b-values, which accrue from fetal motion. Previous techniques for motion compensation include post-acquisition motion compensation based on image registration to bring the volumes acquired at different b-values into the same physical coordinate space before fitting a signal decay model \cite{guyader2015influence,mazaheri2012motion}. However, each b-value image has a different contrast. As a result, independent registration of different b-value images to a b = 0 s/mm$^2$ image may result in suboptimal registration, especially for high b-value images where the signal is significantly attenuated and the signal-to-noise ratio is low \cite{Kurugol2017Motion-robustEstimation}.

\section{Method}
We hypothesize that simultaneous qDWI analysis and fetal motion compensation will result in improved quantification of the ADC, which will correlate better with fetal lung maturity compared to qDWI analysis without compensating for motion and to motion compensated approach which does not account for the signal decay model fit quality.

We define the simultaneous qDWI analysis and motion compensation problem as follows:
\begin{align}
    \widehat{\Phi},\widehat{ADC} = \argmin_{\Phi,ADC} \sum_{i \in B}\|\phi_i \circ S_i-S_0\exp(-b_i\cdot ADC)\|^2 
    \label{eq:problem_formulation_base}
\end{align}
where $\Phi=\{\phi\}_{i \in B}$ is the set of transformations that align the different b-value images to the images predicted by the model, and $B$ is the number of b-values used during the DWI acquisition.

Taking an unsupervised DNN-based perspective, the optimization problem seeks to find the DNN weights that minimize the following:
\begin{align}
    \widehat{\Theta} = \argmin_{\Theta} \sum_{i \in B}\|f_{\Theta}(\{b_i,S_i\}_{i \in B})[0] \circ S_i-S_0\exp(-b_i\cdot f_{\Theta}(\{b_i,S_i\}_{i \in B})[1])\|^2 
    \label{eq:problem_formulation_deep}
\end{align}
where $\Theta$ are the DNN parameters to be optimized,  $f_{\Theta}(\{b_i,S_i\}_{i=0}^{B})[0]$ is the first output of the  DNN model, represents the set of transformations $\Phi$ between the different b-value images $\{S_i\}_{i \in B}$, and $f_{\Theta}(\{b_i,S_i\}_{i \in B})[1])$ is the second output of the DNN model represents the predicted ADC.

\subsection{DNN architecture}
\begin{figure}[t]
	\centering
    \includegraphics[width=\textwidth]{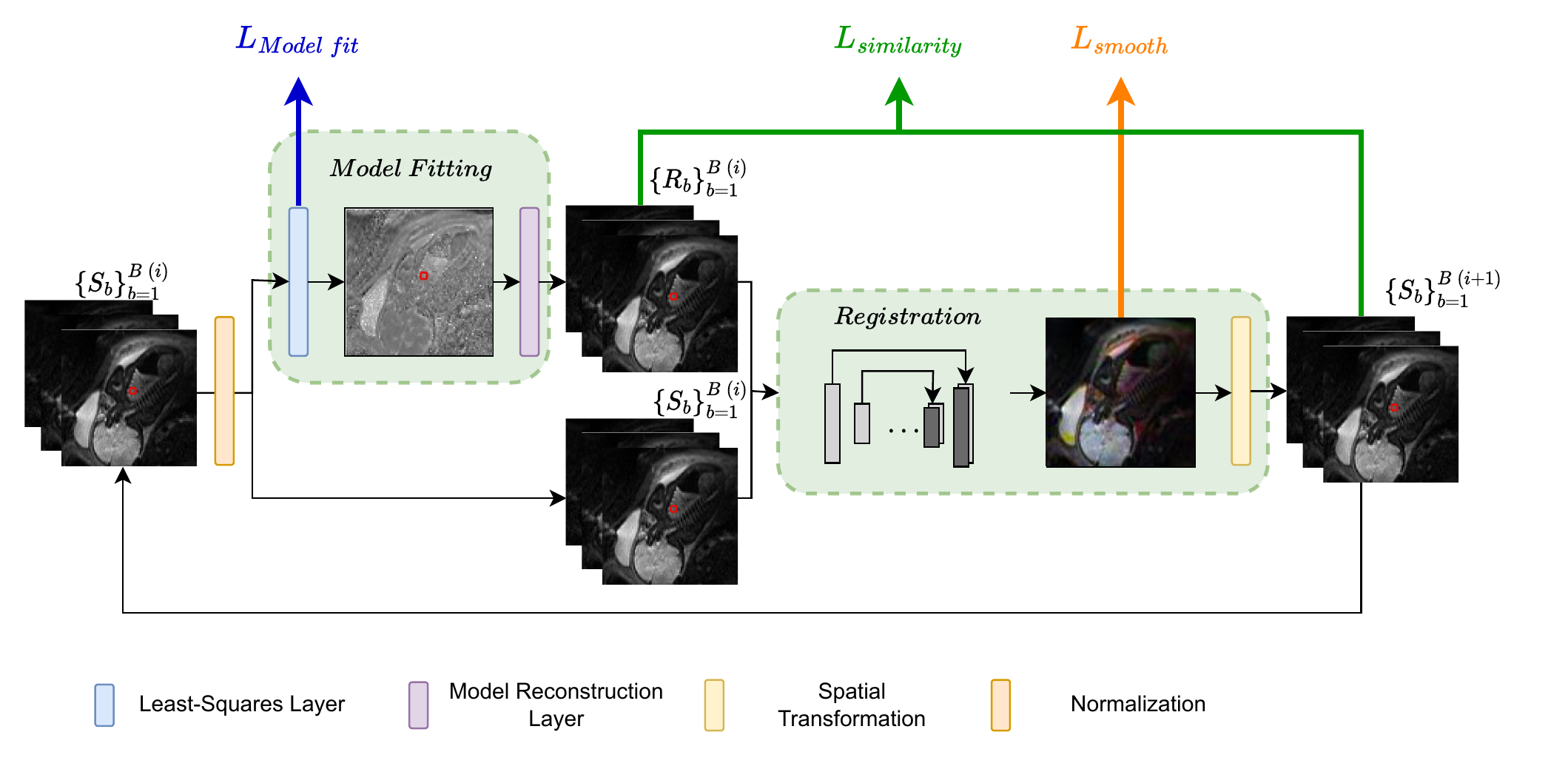}
    \caption{Overall architecture of our proposed framework.}
	\label{Arictecture}
\end{figure}

Fig.~\ref{Arictecture} presents our qDWI-Morph architecture used for the optimization of eq.~\ref{eq:problem_formulation_deep}. The qDWI-Morph model is composed of two sub-networks: a qDWI model fitting sub-network and an image registration sub-network.
Our method is iterative, where the motion-corrected images from the previous iteration enter as input to the next iteration until convergence. We normalize the input in each iteration $k$ by the maximal value of the signal at the previous iteration $S_0^{(k-1)}$.
The input to the model at each iteration $k$ is a set of 3D images from the same patient, each scanned with a different b-value along with their corresponding b-values $\left \{b_i,S_i \right \}_{i \in B}^{(k)}$. 
The output of our model is composed of the set of motion-corrected images forced pixel-wise to the mono-exponential model (Eq.~\ref{eq:MonoModel}). 

\subsubsection{Quantitative DWI Model Fitting Sub-network} first layer is a LLS layer for quantifying the ADC at each pixel using Eq.~\ref{eq:LLS}. The second layer is a reconstruction layer that generates a new set of 3D images, $\left \{ R_b \right \}_{b \in B}^{(k)}$, by calculating the model pixel-wise using Eq.~\ref{eq:MonoModel}.
The reconstructed set, $\left \{ R_b \right \}_{b \in B}^{(k)}$, is retreated as fixed images and is an input for the second sub-network, the registration network.

\subsubsection{Registration Sub-network} is based on the publicly available deep learning framework for deformable medical image registration, VoxelMorph \cite{balakrishnan2019voxelmorph}. The moving images are $\left \{ S_b \right \}_{b \in B}^{(k)}$, and the fixed one are  $\left \{ R_b \right \}_{b \in B}^{(k)}$. The output of the VoxelMorph network is a registration field, $\phi$, that can be applied to the moving images to get the next motion-compensated images, $\left \{ S_b \right \}_{b \in B}^{(k+1)} $. The registration is done between corresponding b-values images from the moving and fixed images. For example, the moving image $S_0$ will be registered to the fixed image $R_0$. 
This way, we overcome the differences in contrast between b-values and use prior knowledge that the resulting compensated image needs to follow the mono-exponential model.

\subsection{Convergence Criteria}
In each iteration, we calculate the averaged ADC of the lung at the given region of interest (ROI). We quantify the ADC by the IRLS algorithm from Eq.~\ref{eq:IRLS}.
We define convergence of the model after five iterations without change in the calculated ADC.
We leverage our prior knowledge of the model (Eq.~\ref{eq:MonoModel}) to choose the best iteration as the iteration with the highest $R^2$ from the IRLS fitting.

\subsection{Bio-physically-informed Loss Function}
Our bio-physically-informed loss function consists of 3 terms.
\begin{align}
L = L_{similarity} + \alpha_1 L_{smooth} +\alpha_2 L_{Model\: fit} 
\end{align}

\subsubsection{L$_{similarity}$} is a $L_1$ loss applied to corresponding b-values images from fixed and wrapped moving images, averaged over the $B$ b-values images:
 \begin{align}
 L(R^{(k)},S^{(k)}\circ \phi_i^{(k)}) = \frac{1}{B} \frac{1}{|\Omega|}\sum_{i\in B}\sum_{p\in \Omega}|R_i^{(k)}(p)-[S_i^{(k)}\circ \phi_i^{(k)}](p)|
 \end{align}
 where $\Omega$ is the image dimension. Minimizing $L_{similarity}$ will encourage  $\left \{ S_i \right \}_{b \in B}^ {(k)} \circ \phi_i^{(k)}$ to approximate  $\left \{ R_i \right \}_{b \in B}^{(k)}$, but may generate a non-smooth $\phi$ that is not physically realistic \cite{balakrishnan2019voxelmorph}.
  
 \subsubsection{L$_{smooth}$} is a regularization term introduced by Balakrishnan et al. \cite{balakrishnan2019voxelmorph}.
 This loss term encourages a smooth displacement field $\phi$ using a diffusion regularize on the spatial gradients of displacement $u$: 
 \begin{align}
 L_{smooth}(\Phi) = \sum_{p\in\Omega} \left \|\nabla u(p)  \right \|^2
 \end{align}

 \subsubsection{L$_{model\ fit}$} is the loss that responsible for forcing each voxel to act according to the mono-exponential model from Eq.~\ref{eq:MonoModelLog}. This loss is an MSE loss on the least square residual: 
 

 \begin{align}
 L_{model\ fit}(ADC) =  \frac{1}{B}\frac{1}{\left | \Omega^* \right |}\sum_{i\in B}\sum_{p \in \Omega^*} (log(S_0)-b_i\cdot ADC-log(S_i))^2
 \end{align}

Where $\Omega^*$ is a subset of the image, including only voxels from the lung, as we assume the mono-exponential model can describe them, contrary to the background, which doesn't present a mono-exponential decay of the DWI signal.

\section{Experiments}
\subsection{Clinical Data-set}
Legacy fetal DWI data was used in the study. The data acquisition was performed on a Siemens 3T Skyra scanner equipped with an 18-channel body matrix coil. Each patient was scanned with a multi-slice, single shot, echo-planar imaging (EPI) sequence that was used to acquire diffusion-weighted scans of the lungs. The in-plane resolution was 2.5mm × 2.5mm for each study, and the slice thickness was set at 3mm. Echo time (TE) was 60ms, whereas repetition time (TR) ranged from 2s to 4.4s depending on the number of slices required to cover the lungs. Each patient was scanned with 6 different b-values (0, 50, 100, 200, 400, 600 sec/mm$^2$) in axial and coronal planes. A region of interest (ROI) was manually drawn for each case in the right lung \cite{Afacan2016FetalAge}. 
We cropped each image to a shape: $96 \times 96 \times 16$ and normalized it by the maximal value at $S_0$. 
The data includes 38 cases with a range of minor misalignment between the different b-values image volumes.  


\subsection{Experiment Goals}
The objectives of the experiment are:
1) To analyze the effect of quantitative DWI motion-compensation on the correlation between the ADC parameter and GA. 2) To analyze the contribution of our proposed model fitting quality loss in addition to the registration loss in improving the observed correlation between the ADC parameter and GA.

\subsection{Experimental Setup}

Our baseline method is ADC quantification without motion compensation. We compared the baseline method to our suggested method with: 1) Registration loss solely, and 2) Our hybrid loss combining both registration loss and a model fitting quality loss.  
For each of the three methods, we quantified the averaged ADC in the right lung of each subject by IRLS algorithm. We used an exponential saturation model suggested by Afacan et al. \cite{Afacan2016FetalAge} described in Eq.~\ref{eq:GA_ADC} above.


\subsection{Implementation details}
We implemented our models on PyCharm 2021.2.3, Python 3.9.12 with PyTorch 1.11.0.

We applied our suggested methods with a batch size of one, meaning that each batch is data from one patient with size: $n_b \times n_x \times n_y \times n_z $, where $n_b$ is the number of b-values used for scanning the patient, and $n_x \times n_y \times n_z$ is the image shape. 
We trained the networks in two epochs so that the earlier cases will be contributed from the later ones. We experimented the affect of training in a wide range of epochs and found that the network converged after the second epoch, and there was no further improvement after that.
We used an Adam optimizer with an initial learning rate of 10-4. If the total loss increases compared to the previous iteration, we reduce the learning rate by 10. We limited the number of iterations for each case to 50.
The hyper-parameter we used are: $\alpha_1 = 0.01$, $\alpha_2 = 1000$. We choose $\alpha_2$ by a search-grid of the best $\alpha_2$ in terms of correlation between ADC and GA. For testing the qDWI-Morph method without fit model loss term, we set $\alpha_2 = 0$. 

\section{Results}



\begin{figure}[t]
     \centering
     \begin{subfigure}[b]{0.31\textwidth}
         \centering
         \includegraphics[width=\textwidth]{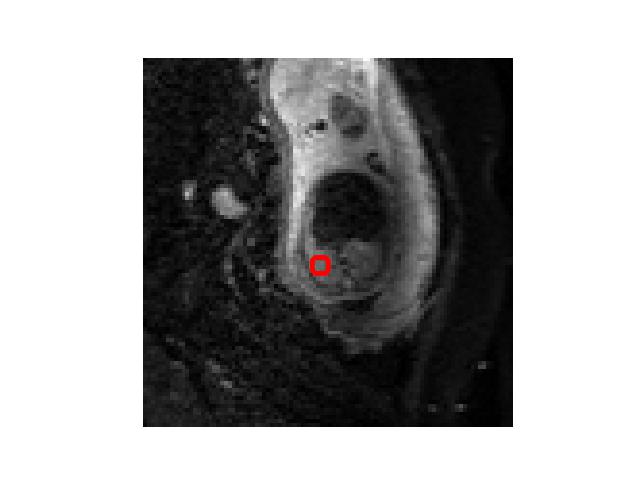}
         \caption{}
         \label{fig:Original b0}
     \end{subfigure}
     \hfill
     \begin{subfigure}[b]{0.31\textwidth}
         \centering
         \includegraphics[width=\textwidth]{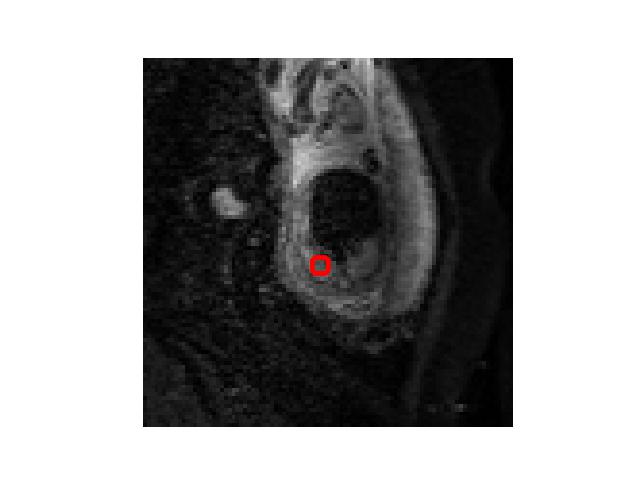}
         \caption{}
         \label{fig:Original b1}
     \end{subfigure}
     \hfill
     \begin{subfigure}[b]{0.31\textwidth}
         \centering
         \includegraphics[width=\textwidth]{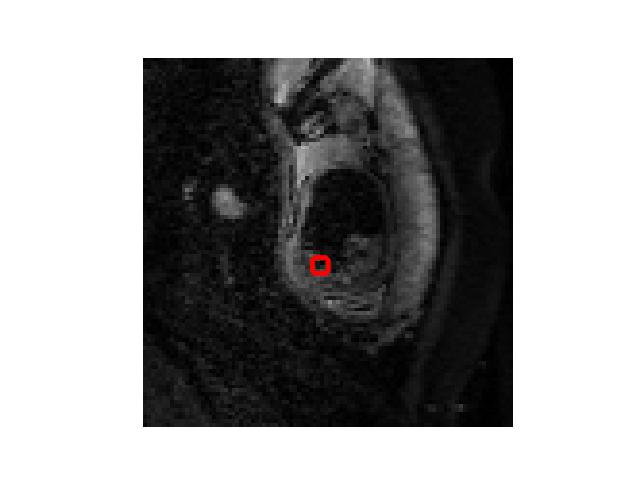}
         \caption{}
         \label{fig:Original b2}
     \end{subfigure}
     \hfill
          \begin{subfigure}[b]{0.31\textwidth}
         \centering
         \includegraphics[width=\textwidth]{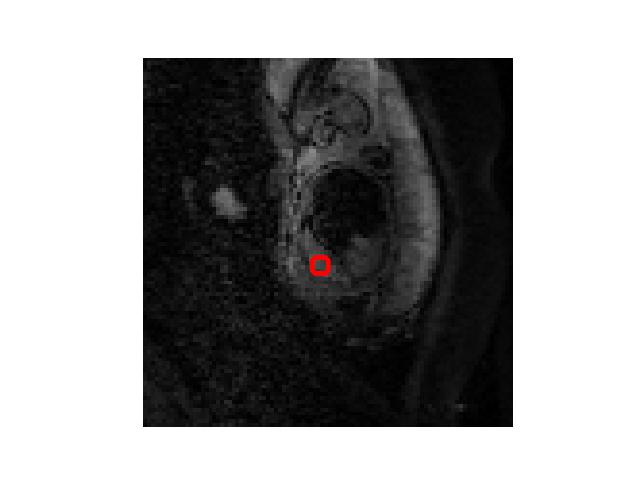}
         \caption{}
         \label{fig:Original b3}
     \end{subfigure}
     \hfill
          \begin{subfigure}[b]{0.31\textwidth}
         \centering
         \includegraphics[width=\textwidth]{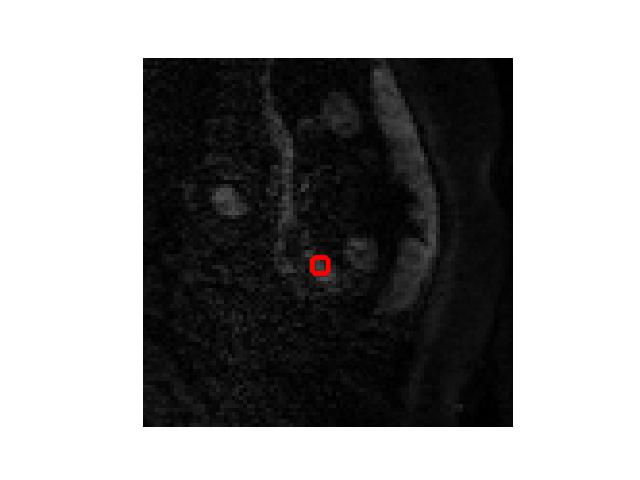}
         \caption{}
         \label{fig:Original b4}
     \end{subfigure}
     \hfill
          \begin{subfigure}[b]{0.31\textwidth}
         \centering
         \includegraphics[width=\textwidth]{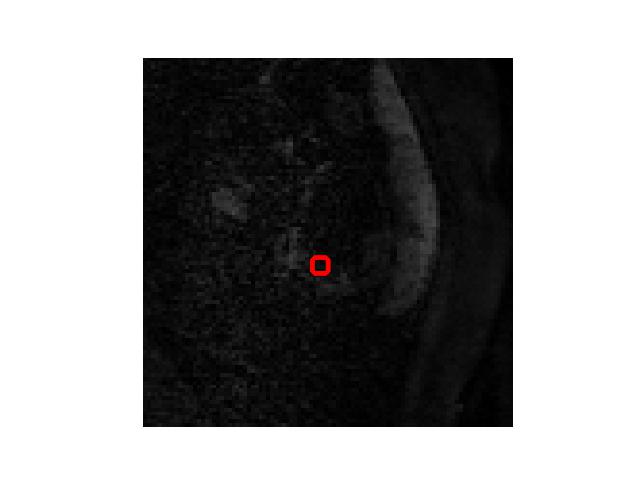}
         \caption{}
         \label{fig:Original b5}
     \end{subfigure}
        \caption{Sample case with ROI. Top row left to right: b = 0, 50, 100 s/mm$^2$. Bottom row left to right: b-value = 200, 400, 600 s/mm$^2$.}
        \label{fig:Case_691_Original}
\end{figure}
\begin{figure}[t]
	\centering
    \includegraphics[height=9cm]{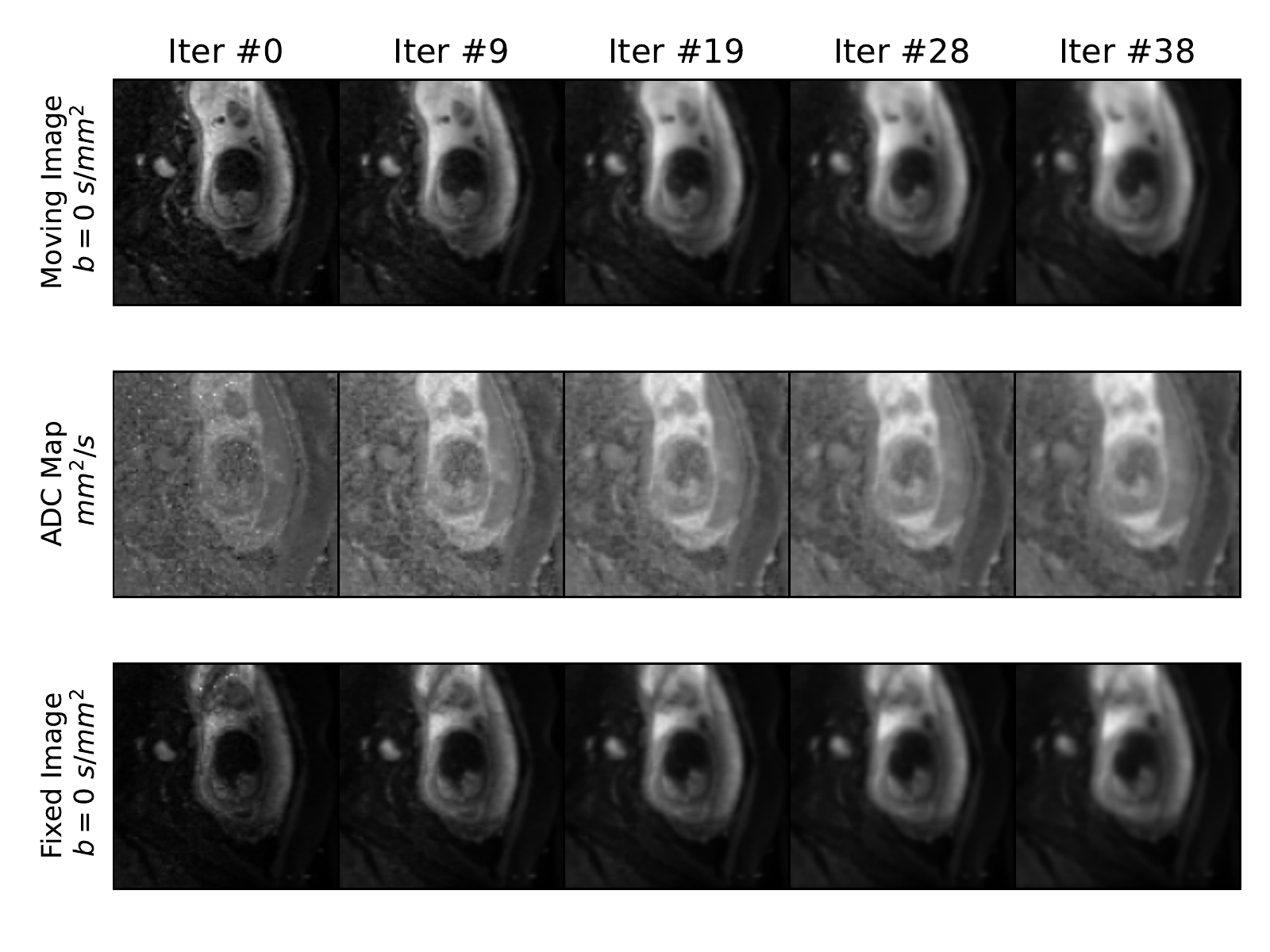}
    \caption{The process done by our method, qDWI-Morph, demonstrated over time. Top row: the motion-compensation b=0 s/mm$^2$ images. Second row: ADC map result from the LLS layer. Bottom row: the model-reconstructed b=0 s/mm$^2$ images. 
}
	\label{fig:Case_691_Summery}
\end{figure}

Fig.~\ref{fig:Case_691_Original} presents an example of DWI data before accounting for motion by the registration. Fig.~\ref{fig:Case_691_Summery} presents our iterative motion-compensation and model estimation. 
The top row is the motion-compensation b=0 s/mm$^2$ images from different iterations, where the top-left image is the input image. The second row is the ADC map result from the LLS layer, and the bottom row is the corresponded reconstructed b=0 s/mm$^2$ image. The right column is the convergence iteration. The ADC map (middle row) becomes sharper and more clinically-meaningful as the motion is correct during the iterations. However, the final motion-compensated b-value images become more blurry as a result of the registration wrapping process by the linear interpolation. 

\begin{figure}[t]
     \hspace*{\fill}
     \begin{subfigure}[b]{0.32\textwidth}
         \centering
         \includegraphics[width=\textwidth]{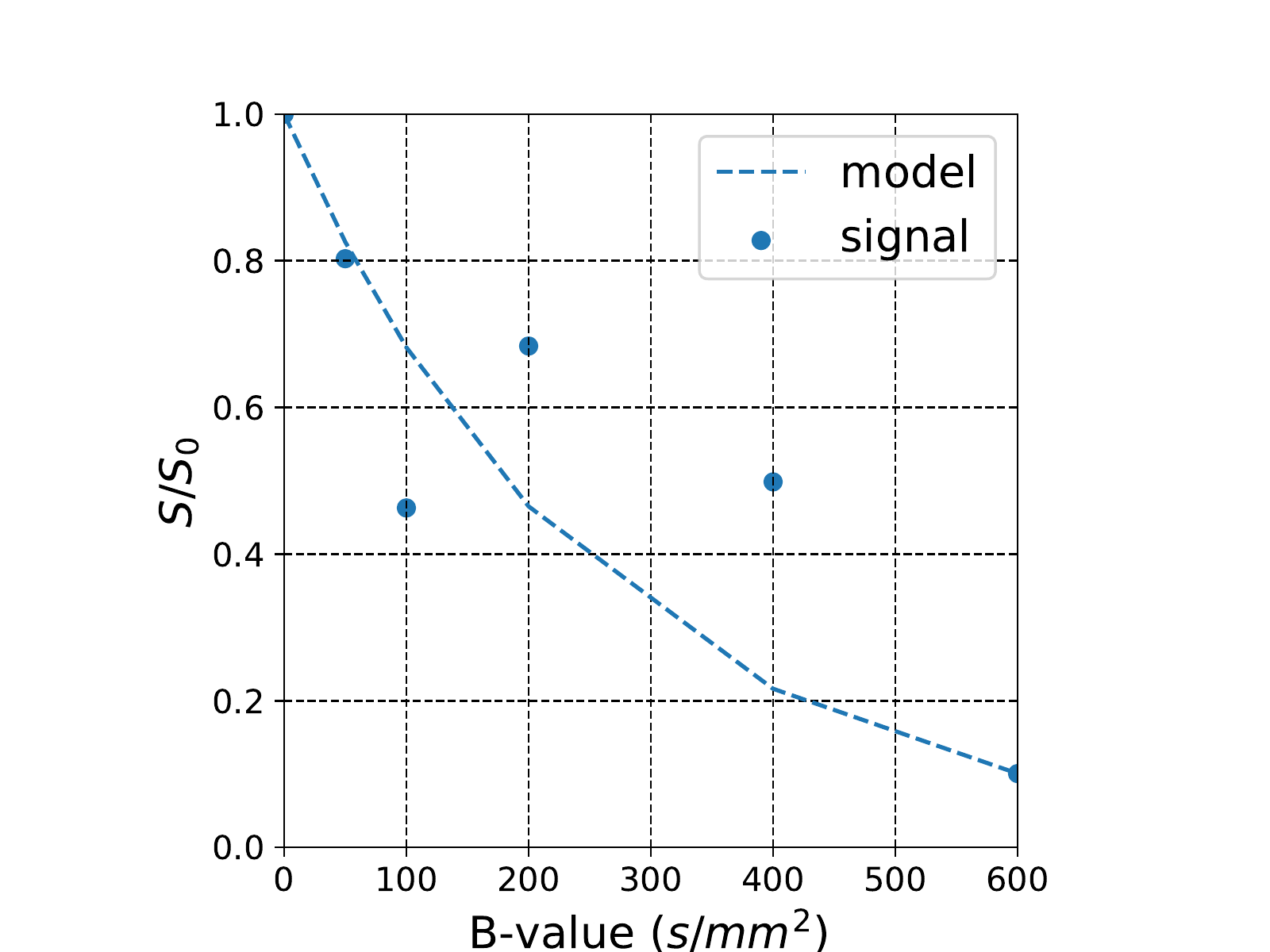}
         \caption{}
         \label{fig:ROI iter 0}
     \end{subfigure}
     \hfill
     \begin{subfigure}[b]{0.32\textwidth}
         \centering
         \includegraphics[width=\textwidth]{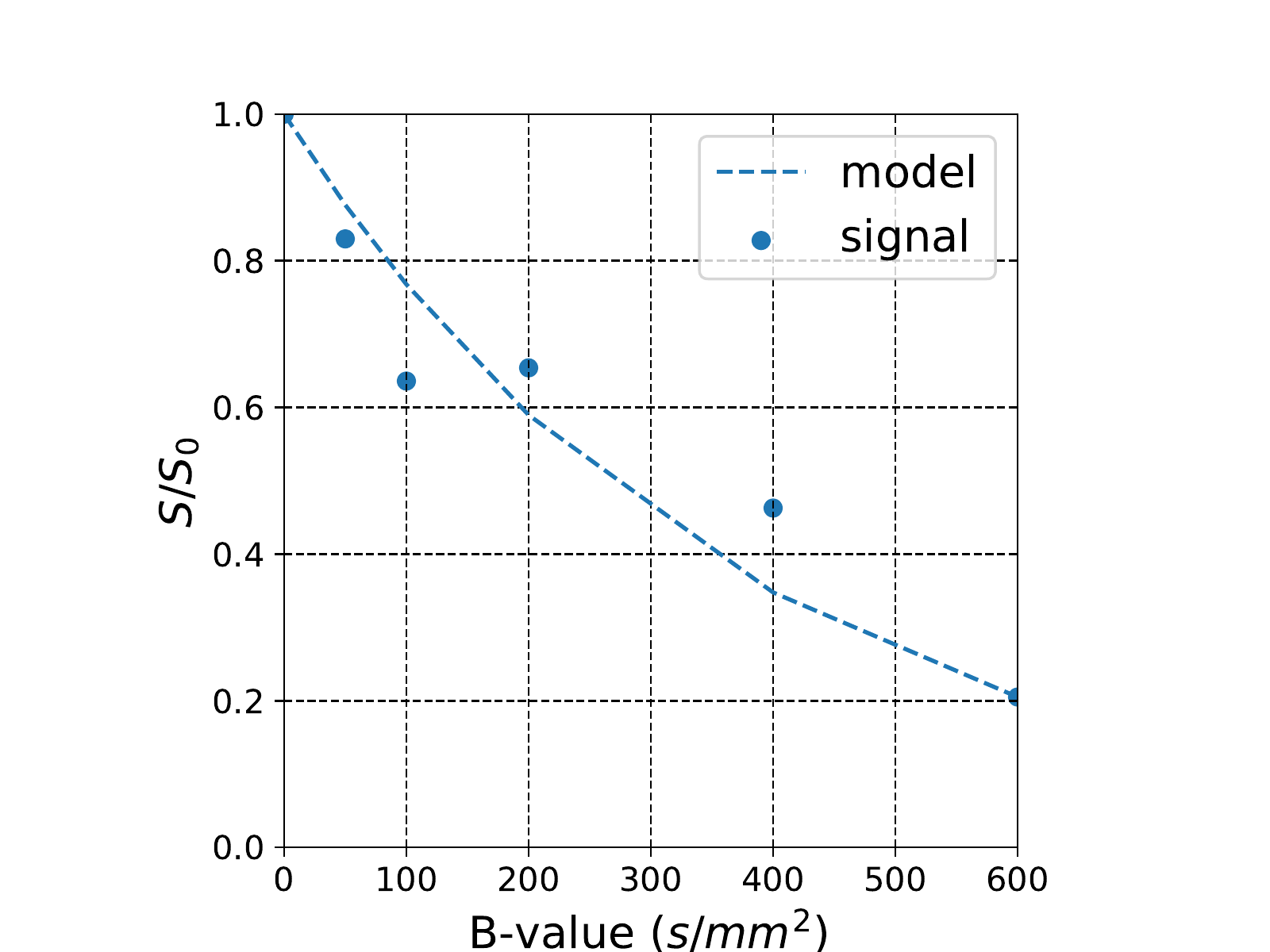}
         \caption{}
         \label{fig:ROI iter 19}
     \end{subfigure}
     \hfill
     \begin{subfigure}[b]{0.32\textwidth}
         \centering
         \includegraphics[width=\textwidth]{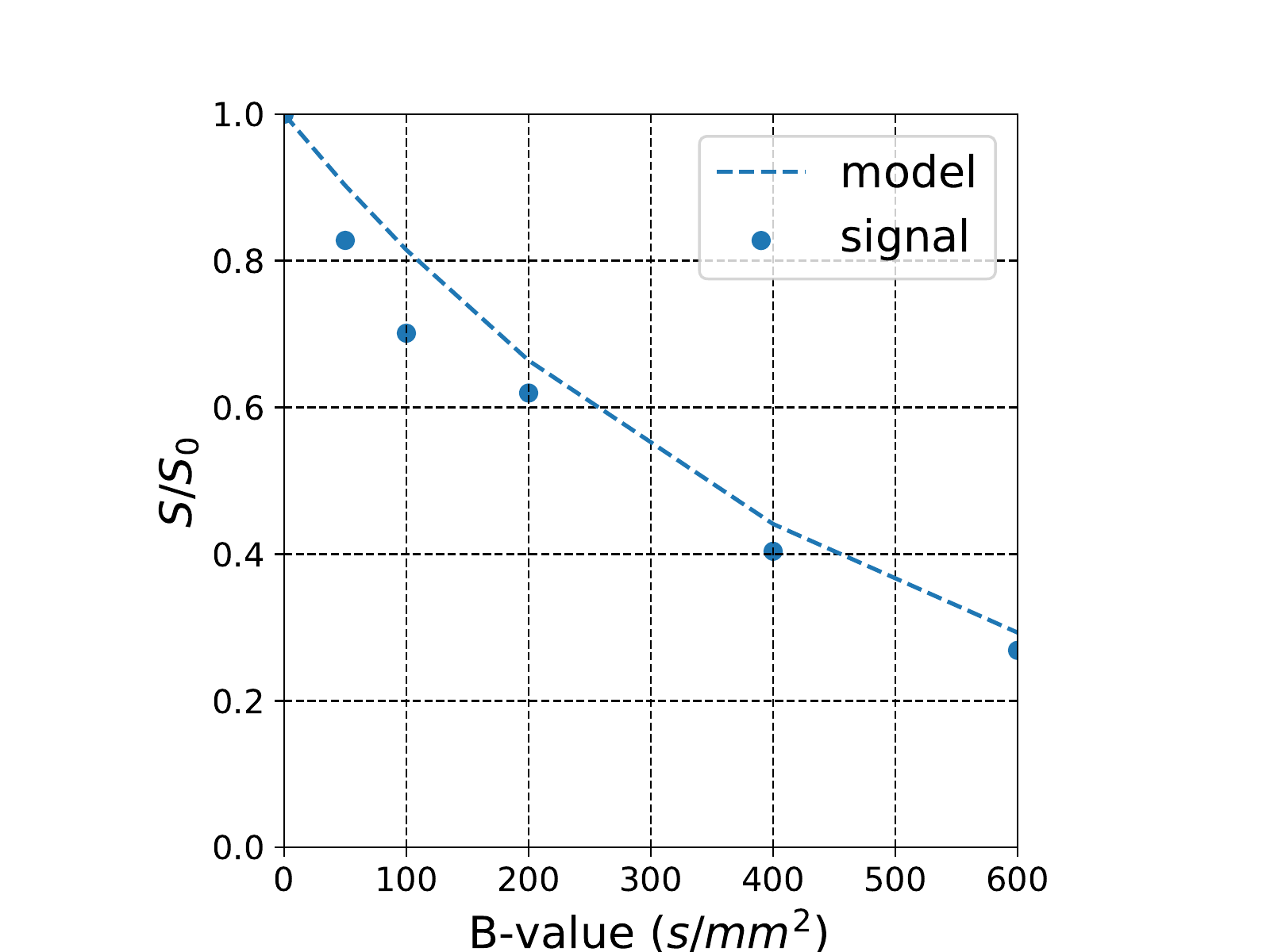}
         \caption{}
         \label{fig:ROI iter 38}
     \end{subfigure}
     \hspace*{\fill}
        \caption{The averaged signal in the ROI for one case, as a function of b-value at different iterations, with a curve fit to the mono-exponential model: (a) Original data, (b) iteration 19, and (c) iteration 38.}
        \label{fig:ROI 867}
    
\end{figure}

Fig.~\ref{fig:ROI 867} demonstrates the averaged signal in the ROI as a function of the b-value at different iterations. The lines are the corresponding fit to Eq.~\ref{eq:MonoModel}. At the last iteration, after the motion compensation process (\ref{fig:ROI iter 38}), the dots behave according to the expected model and are less scattered.

Fig.~\ref{fig:ADC_GA_qDWIMorph} shows the ADC parameter plotted against GA for all three methods. The data is color-coded by the degree of conformity to the mono-exponential model from Eq.~\ref{eq:MonoModel} in terms of $R^2$. The Darker the dot, the higher $R^2$. Our approach achieved the best correlation ($R^2=0.32$) between the ADC and the GA compared to both motion compensation by our qDWI-Morph without adding the model fit loss ($R^2=0.28$) and to baseline approach without motion compensation ($R^2=0.13$).

\begin{figure}[t]
     \centering
     \begin{subfigure}[b]{0.325\textwidth}
         \centering
         \includegraphics[width=\textwidth]{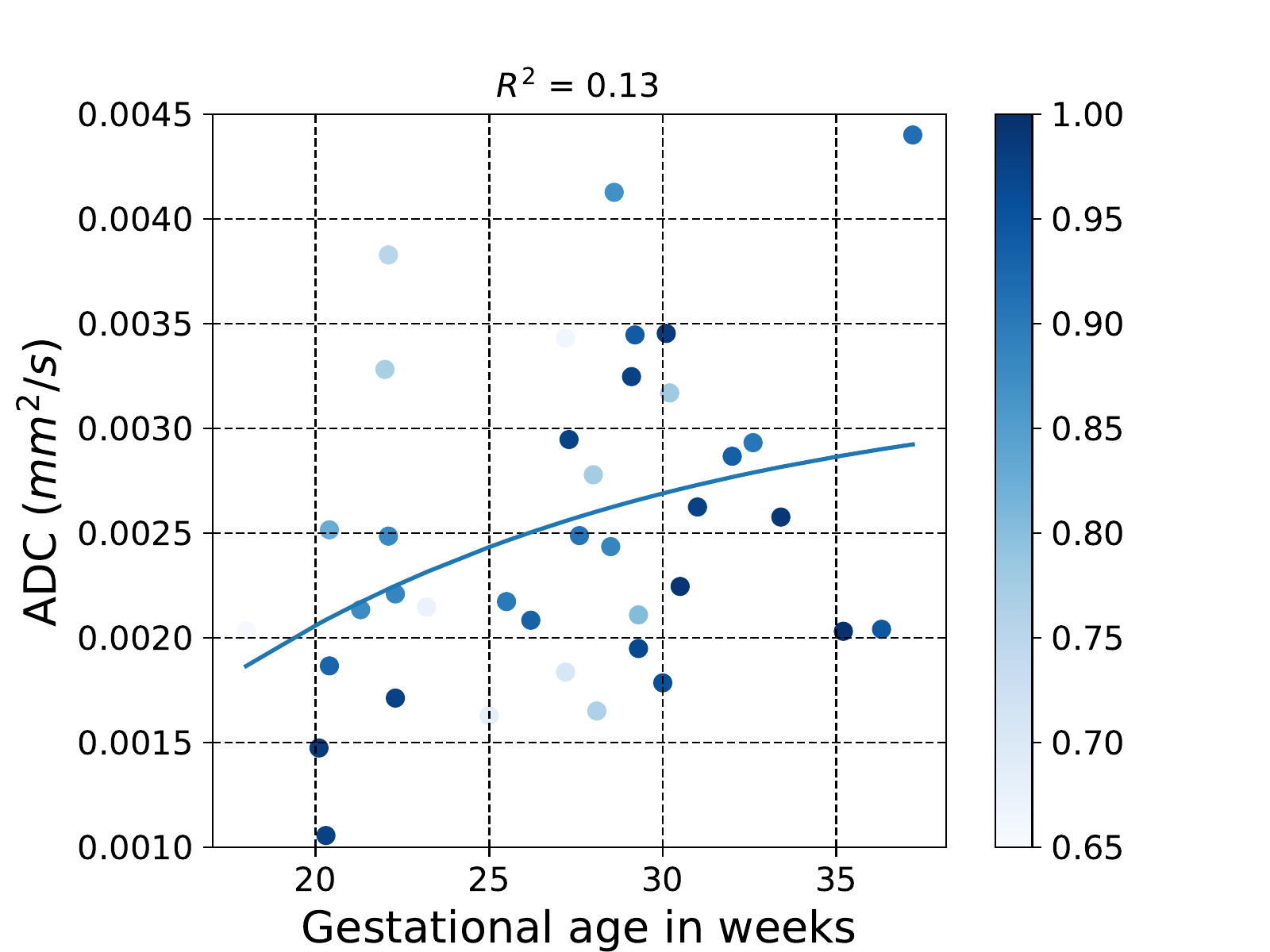}
         \caption{}
         \label{fig:ADC_GA_NoMotion}
     \end{subfigure}
     \hfill
     \begin{subfigure}[b]{0.325\textwidth}
         \centering
         \includegraphics[width=\textwidth]{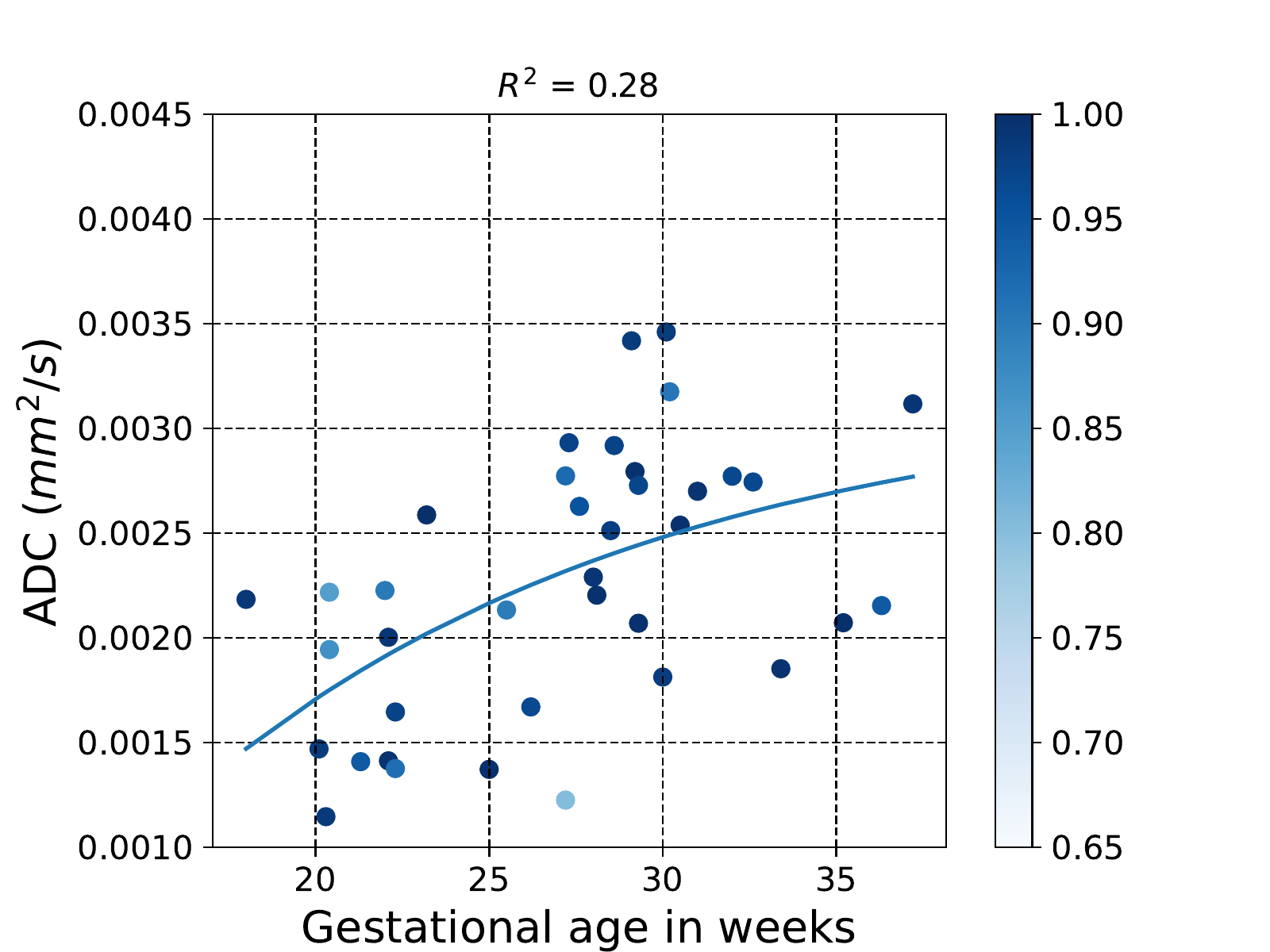}
         \caption{}
         \label{fig:ADC_GA_NoFit}
     \end{subfigure}
     \hfill
     \begin{subfigure}[b]{0.325\textwidth}
         \centering
         \includegraphics[width=\textwidth]{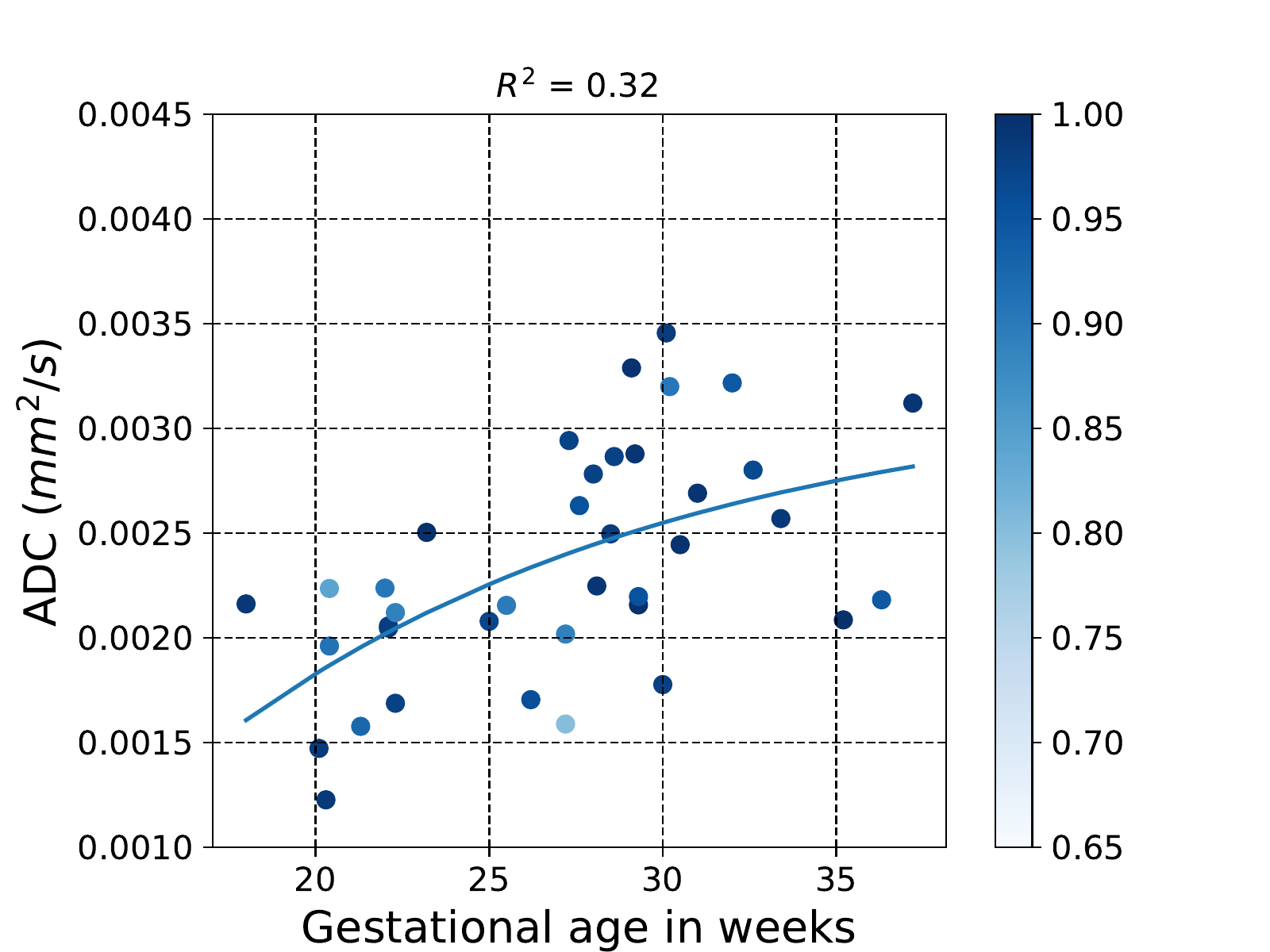}
         \caption{}
         \label{fig:ADC_GA_qDWIMorph}
     \end{subfigure}
     \caption{Averaged ADC in the right lung vs. GA with the three methods: (a) Baseline; W.O motion compensation. (b) qDWI-Morph W.O model fit loss. (c) qDWI-Morph.The data is color-coded by the degree of conformity to the mono-exponential model from Eq.~\ref{eq:MonoModel} in terms of $R^2$.}
        \label{fig:ADC_GA_qDWIMorph}
\end{figure}

\setlength{\tabcolsep}{2pt}
\begin{table}[tp]
\begin{center}
\caption{The correlation of the averaged ADC in the right lung vs. GA by the three methods: W.O M.C; the baseline method, ADC quantification without motion compensation. W.O F.L; qDWI-Morph without model fit loss, and our method: q-DWI-Morph.}
\label{table:GA_ADC_corr}
\begin{tabular}{lcccc}
\hline\noalign{\smallskip}
Method & $R^2$ & \multicolumn{1}{p{2cm}}{\centering A $\times 10^{-3}$ \\ $(mm^2/s)$} 
&  \multicolumn{1}{p{2cm}}{\centering B $\times 10^{-3}$ \\ $(mm^2/s)$} & C\\
\noalign{\smallskip}
\hline
\noalign{\smallskip}
W.O. M.C & 0.13 & 3.2 & 0.005 & 0.07 \\
W.O. F.L  & 0.28 & 3.17 & 0.006 & 0.07 \\
qDWI-Morph  & {\bf 0.32} & 3.19 & 0.006 & 0.07 \\

\hline
\end{tabular}
\end{center}
\end{table}
\setlength{\tabcolsep}{1.4pt}

Table~\ref{table:GA_ADC_corr} summarizes the fitted ADC saturation model (Eq.~\ref{eq:GA_ADC}) using the different approaches along with the correlation between the ADC and GA in terms of $R^2$. The fitted model parameters are bout the same for all methods. However, our qDWI-Morph achieved the best correlation between the ADC and the GA, since the average ADC vs. GA is less sparse, as seen in Fig.~\ref{fig:ADC_GA_qDWIMorph}.

\section{Conclusions}
We introduced qDWI-Morph, an unsupervised deep-neural-networks approach for Motion-compensated quantitative Diffusion-Weighted MRI analysis with application in fetal lung maturity assessment. Our model coupled a registration sub-network with a model fitting sub-network to simultaneously estimate the motion between the different b-value images and the signal decay model parameters. The optimization of the model weights is driven by minimizing a bio-physically-informed loss representing both the registration loss and the fit quality loss. The integration of the fit quality loss encourages the model to produce deformation fields that will lead to bio-physically expected behavior of the signal along the b-value axis, in addition to the standard registration loss, which encourages similarity and smoothness within the deformation field of each b-value image. Our experiments demonstrated an added-value of adding the model-fitting loss in addition to the registration loss for fetal lung maturity assessment. The proposed approach can potentially improve our ability to quantify DWI signal decay model parameter in cases with motion.

%
%
\bibliographystyle{splncs04}
\bibliography{egbib}

\end{document}